\documentclass{article}

\usepackage{microtype}
\usepackage{graphicx}
\usepackage{subfigure}
\usepackage{amssymb}
\usepackage{amsmath}
\usepackage{pifont}
\usepackage{booktabs} % for professional tables

\usepackage{hyperref}

\newcommand{\cmark}{\ding{51}}%
\newcommand{\xmark}{\ding{55}}%

\usepackage[accepted]{icml2020}

\icmltitlerunning{Sparse-Push: Communication and Energy Efficient Decentralized \& Distributed Learning }
\begin{document}

\twocolumn[
\icmltitle{Sparse-Push: Communication- \& Energy-Efficient Decentralized Distributed Learning 
over Directed \& Time-Varying Graphs with non-IID Datasets}

% It is OKAY to include author information, even for blind
% submissions: the style file will automatically remove it for you
% unless you've provided the [accepted] option to the icml2020
% package.

% List of affiliations: The first argument should be a (short)
% identifier you will use later to specify author affiliations
% Academic affiliations should list Department, University, City, Region, Country
% Industry affiliations should list Company, City, Region, Country

% You can specify symbols, otherwise they are numbered in order.
% Ideally, you should not use this facility. Affiliations will be numbered
% in order of appearance and this is the preferred way.
\icmlsetsymbol{equal}{*}

\begin{icmlauthorlist}
\icmlauthor{Sai Aparna Aketi}{pu}
\icmlauthor{Amandeep Singh}{imec}
\icmlauthor{Jan Rabaey}{berkeley,imec}
\end{icmlauthorlist}

\icmlaffiliation{pu}{Department of Electrical and Computer Engineering, Purdue University, West Lafayette, USA. }
\icmlaffiliation{imec}{IMEC, Berkeley, USA. }
\icmlaffiliation{berkeley}{BWRC, University of California Berkeley, Berkeley, CA 94704, USA}

\icmlcorrespondingauthor{Sai Aparna Aketi}{saketi@purdue.edu}

% You may provide any keywords that you
% find helpful for describing your paper; these are used to populate
% the "keywords" metadata in the PDF but will not be shown in the document
\icmlkeywords{Communication compression, communication scheduler, decentralized training, directed graphs, distributed machine learning, gossip averaging, non-IID data, quantization, sparsification,  time-varying graphs}

\vskip 0.3in
]

% this must go after the closing bracket ] following \twocolumn[ ...

% This command actually creates the footnote in the first column
% listing the affiliations and the copyright notice.
% The command takes one argument, which is text to display at the start of the footnote.
% The \icmlEqualContribution command is standard text for equal contribution.
% Remove it (just {}) if you do not need this facility.

\printAffiliationsAndNotice{}  % leave blank if no need to mention equal contribution
%\printAffiliationsAndNotice{\icmlEqualContribution} % otherwise use the standard text.

\begin{abstract}
Current deep learning (DL) systems rely on a centralized computing paradigm which limits the amount of available training data, increases system latency and adds privacy \& security constraints. On-device learning, enabled by decentralized and distributed training of DL models over peer-to-peer wirelessly connected edge devices, not only alleviate the above limitations but also enable next-gen applications that need DL models to continuously interact and learn from their environment. However, this necessitates development of novel training algorithms that train DL models over time-varying \& directed peer-to-peer graph structures while minimizing the amount of communication between the devices and also being resilient to non-IID data distributions. In this work we propose, \emph{Sparse-Push}, a communication efficient decentralized distributed training algorithm that supports training over peer-to-peer, directed and time-varying graph topologies. 
The proposed algorithm enables 466$\times$ reduction in communication with only $1\%$ degradation in performance when training various DL models such as ResNet-20 and VGG11 over CIFAR-10 dataset. Further, we demonstrate how communication compression can lead to significant performance degradation in-case of non-IID datasets, and propose \emph{Skew-Compensated Sparse Push} algorithm that recovers this performance drop while maintaining  similar levels of communication compression.
\end{abstract}

\section{Introduction}
\label{intro}
Deep learning and deep reinforcement learning  models have achieved remarkable success on a variety of tasks such as image classification and segmentation, object detection, robotic perception, manipulation and navigation, etc. This large success of deep learning and deep reinforcement learning models can largely be attributed to three key things: (a) Improvements in deep learning architectures and learning algorithms, (b) Presence of large data sets, and (c) An exponential increase in available compute capabilities. However, almost all currently used learning approaches rely heavily on a centralized computing paradigm where-in we gather a large amount of data from numerous edge devices, transfer this tremendous data to data-centers or cloud machines, and subsequently train large-scale deep learning models using enormous computational power. Such centralized computing solutions, while successful for few industrial use cases, consume large amount of power and network bandwidth to transfer data from edge devices to data centers. In fact, the huge amount of power needed to communicate raw data from edge to cloud, fundamentally limits the maximum amount of data we can use for training our deep learning systems. For instance, per CISCO's latest estimates, about 857 Zettabytes of data is created annually, of which only 5.9 Zettabytes is stored on devices, and only 1.3 Zettabytes is actually sent to the cloud. Thus, our current learning solutions only use about 1\% of the total data. Further, perhaps more importantly, current centralized computing paradigm does not work well for next generation low-latency applications such as embodied AI, AR/VR, and various robotic applications for unstructured world where the edge devices or robotic agents need to continuously interact with and learn from the environment with an extremely tight latency budget of a fraction of a millisecond. Moreover, the centralized computing paradigm also brings about a number of privacy, and data security constraints as well.  

A potential solution that helps overcome the above challenges is to allow distributed and decentralized deep learning over a graph of peer-to-peer wireless edge devices each embedded with its own on-device learning capability. On-device learning, coupled with distributed and decentralized training, helps avoid transferring raw data from edge devices to a centralized cloud while allowing the edge devices to make maximum use of available data. Further, it significantly reduces latency, data privacy and security issues. However, decentralized and distributed learning over wirelessly connected edge devices devices brings about its unique set of challenges. In particular, edge devices are severely energy constrained and need to judiciously choose \textit{``what and when to communicate"}, such that they can jointly maximize the task performance while minimizing the cost of communication amongst the peers. It may be noted that learning just from the data gathered by a single device itself without communicating amongst the peers will lead to poor task performance as each edge device only gets to observe a small subset of data. For instance, in an image classification task, a particular edge device may only observe few categories and it relies on communication with its peers to learn how to correctly classify other categories. Effectively, the data observed by each edge device is distributed in a non-IID manner. It may also be noted that unlike distributed optimization in data center settings where the connectivity graph is largely static, the connectivity graph for decentralized and distributed learning over wirelessly connected edge devices is largely time varying. Furthermore, in such a setting, each edge device may either broadcast information to all nearby devices, or it could selectively send information to a carefully chosen subset of edge devices. In either case, each edge device can only control which other edge devices it can send information to, but it does not know apriori the devices from whom it will receive information from. This makes the connectivity graph amongst the edge devices directed and asymmetric. It may be noted that this violates the commonly held assumption of ``doubly-stochastic connectivity graph" \cite{top} used in majority of current decentralized distributed optimization algorithms. 

Based on the above constraints, we cast our problem as: \textit{Given a set of 'N' nodes connected in a peer-to-peer time-varying and directed graph topology, can we design a training algorithm that helps learn a deep learning model to achieve equivalent task performance as in a conventional centralized learning setting while minimizing  the amount of communication, and simultaneously allowing for non-IID data distribution amongst the nodes?} 

In this paper, we propose two novel algorithms \textit{Sparse-Push (SP)} and \emph{Skew-Compensated Sparse Push (SCSP)} that help achieve all the above goals simultaneously. 
To the best of our knowledge, this is the first work that minimizes the amount of communication in decentralized and distributed training setting, taking into account directed and time-varying graph structures while also minimizing the degradation caused due to non-IID data distribution. We test the performance of our algorithm on various supervised image classification tasks using ResNet and VGG architectures over CIFAR-10 and CIFAR-100 datasets. We demonstrate $466\times$ reduction in communication over a time-varying directed ring graph topology with $1\%$ degradation in performance, or $~0.8\%$ degradation with $64\times$ reduction in communication. Further, we highlight the interplay between non-IID data distribution and communication compression and demonstrate how communication compression can lead to more than $7\%$ performance degradation in-case of non-IID data partitions. The proposed SCSP training algorithm, helps recover this performance drop and we report $0.5\% - 2\%$ difference in performance between models trained on IID vs non-IID datasets while maintaining the same level of communication compression.  

The rest of paper is organized as follows. In section 2, we review the current state of the art in decentralized and distributed optimization algorithms and highlight the limitations of current work. Section 3 formally describes our proposed algorithm, \textit{Sparse-Push}. Section 4 describes our experimental setup and demonstrates the results of the algorithm from Section 3. Section 5 explicitly highlights the challenges associated with non-IID data partitions, and proposes a skew-compensated sparse push algorithm that helps overcome performance degradation caused by non-IID datasets. Section 6 briefly describes the hardware implications of our proposed work and highlights how such an approach could help reduce energy consumption for on-device training. Finally section 7 concludes the paper and discusses future work. 

\section{Related Work}
In  recent past, there has been a strong push to design communication efficient decentralized stochastic gradient descent (D-SGD) algorithms for distributed deep learning, especially for data-center settings where the connectivity graph is largely static and doubly stochastic. This section reviews the current state of art in this area. 

\begin{table*}[ht]
\centering
\caption{ \centering Comparison of our work with other baseline algorithms. Please refer to Section 2 for detailed discussion.}
\label{tab:c1}
\begin{small}
\begin{tabular}{|c|c|c|c|c|}
\hline
Method & Peer-to-peer & Communication & Time Varying and  & Support for Non-IID \\
 & Learning & Compression &  Directed Graphs & data distribution\\
\hline
  Skew-Scout \cite{hsieh20a} & \xmark&\cmark & \xmark& \cmark \\
  \hline
   D-PSGD \cite{dsgd} & \cmark&\xmark & \xmark&\xmark\\
 \hline
  $D^2$ \cite{tang2018d2} & \cmark&\xmark & \xmark& \cmark \\
  \hline
   Deep-Squeeze \cite{deepsqueeze} &\cmark & \cmark & \xmark&\xmark\\
   \hline
   CHOCO-SGD \cite{koloskova20} &\cmark & \cmark & \xmark&\xmark\\
 \hline
  SGP \cite{assran19a} & \cmark & \xmark & \cmark & \xmark  \\
  \hline
  Quant-SGP \cite{taheri20a} & \cmark & Low & Directed & \xmark  \\
   & & (Upto $8\times$ only) & graphs only & \\
  \hline
   Skew-compensated Sparse-Push (This Work) & \cmark  & \cmark  & \cmark   & \cmark  \\
  \hline
 \end{tabular}
 \end{small}
\end{table*}

\textit{Decentralized Stochastic Gradient Descent (D-SGD):} Distributed-Parallel SGD (D-PSGD) proposed in \cite{NIPS2017}, was one of the first large-scale experiments that showed the promise of decentralized SGD by demonstrating comparable performance to centralized SGD for deep learning applications. They were the first to show that D-SGD enjoys the same asymptotic convergence as centralized SGD. Subsequently, \cite{adsgd} proposed an asynchronous extension to D-PSGD which allows for asynchronous communication amongst worker nodes. Both the above approaches use gossip averaging to get exact averages at convergence, but only work with static and symmetric doubly stochastic matrices, making them unsuitable for practical directed and time varying graph topologies. In \cite{assran19a}, authors propose stochastic gradient push (SGP), which effectively combines the well known push-sum algorithm \cite{pushsum} together with SGD to enable decentralized SGD over directed and time-varying graph topologies. However, all the above methods require each node to communicate the full model with its neighbours at each communication step, which leads to a huge communication overhead, making them unsuitable for wireless peer-to-peer based edge AI applications. 

\textit{Communication Efficient D-PSGD:} Various methods have been explored to reduce communication overhead in D-PSGD. \cite{dcd} proposed difference compression D-PSGD (DCD-PSGD) and extrapolation compression D-PSGD (ECD-PSGD) algorithms which exchange the compressed difference of local models only. However, these algorithms converge only for a constant compression (unbiased) ratio which significantly limits the maximum possible compression. In \cite{koloskova19a} and \cite{koloskova20}, authors proposed CHOCO-SGD algorithm which allows for arbitrarily high communication compression leveraging an error feedback mechanism while allowing the nodes to communicate the compressed model differences only. However, CHOCO-SGD requires every node to keep track of the model difference between that of its neighbours and its own model. This incurs an additional memory overhead of $O(md)$ where m is the number of edges and d is the model dimension. More importantly, the need to explicitly track the model difference with its neighbours makes it difficult to extend the algorithm to scenarios where the connectivity graph varies with time. Further, \cite{deepsqueeze} proposed an algorithm, Deep-Squeeze, which also implements an error feedback strategy and directly communicates compressed model parameters without the need to explicitly store difference between the local model and its neighbours model. However, all these algorithms assume the connectivity graph to be static and undirected (i.e. the mixing weights to be symmetric and doubly stochastic \cite{top}) and data partitions to be identically and independently distributed (IID).

\textit{Non-IID Datasets and D-SGD:} The effect of non-IID data distribution amongst nodes in D-SGD, especially in the presence of communication compression, is a relatively understudied problem. In \cite{tang2018d2}, authors proposed $D^2$ algorithm to extend D-SGD to non-IID datasets as well. While an interesting approach, the authors achieved limited success and the approach was demonstrated on few basic models, such as LENET only. More importantly, they did not incorporate any communication reduction techniques and based on our experiments $D^2$ fails to converge with non-IID datasets in the presence of communication compression. More recently, \cite{hsieh20a} performed a study on the affect of non-IID data partitions (data skew) on the performance of centralized distributed learning algorithms, i.e. in distributed-SGD in the presence of a centralized server. The authors demonstrated how non-IID data-sets can lead to poor performance and attributed the degradation to, (a) Use of batch-norm, and (b) Amount of gradient compression used while communicating between the nodes to the centralized parameter server. Based on their findings, they proposed to (a) Replace batch-norm with group-norm, (b) Proposed SkewScout algorithm which tries to solve the trade-off between communication compression and performance degradation as an optimization problem. While an interesting approach, repeatedly solving the optimization problem for a time-varying directed graph structure adds a significant compute overhead. Further, it may be noted that unlike decentralized distributed SGD where nodes communicate  models amongst each other, in distributed-SGD with centralized server, the nodes only communicate their gradients amongst each other. Hence, the results from \cite{hsieh20a} do not necessarily carry over directly to decentralized D-SGD settings. 

Table \ref{tab:c1} summarizes the current state of the art and presents our work in context. The recently proposed Quant-SGP algorithm proposed in \cite{taheri20a} is the closest to our work. Quant-SGP extended D-SGD to directed graphs by effectively combining SGP \cite{assran19a} algorithm  with CHOCO-SGD \cite{koloskova20}. However, as mentioned earlier, extending CHOCO-SGD, and hence, Quant-SGP to practical time-varying dynamic graph structures is difficult because of the need for each node to maintain a model for each of its neighbours. Further, Quant-SGP was demonstrated with extremely limited communication compression (upto 8$\times$ compression only) and it does not consider the impact of non-IID data distributions on D-SGD. To the best of our knowledge, our work in this paper is the first to present a unified decentralized and distributed learning algorithm, that works for time-varying and directed graph topologies while supporting large communication compression ratios. Further, we are first to study the effects of non-IID datasets in D-SGD in the presence of communication compression, and propose a simple optimization free methodology to significantly reduce impact of non-IID datasets. 

\section{Sparse-Push (SP) Algorithm}
\label{ds_sgp}
In this section we discuss the proposed \emph{Sparse-Push (SP)} algorithm. Sparse-push algorithm extends deep-squeeze algorithm \cite{deepsqueeze} for communication efficient decentralized and distributed learning to directed and time varying graph topologies by efficiently combining it with push-sum \cite{pushsum}\cite{assran19a} protocol. Please note that the challenges associated with non-IID datasets and communication efficient methods to mitigate them are discussed in section 5. 

\emph{Problem Setup:} Consider $n$ nodes (workers or devices) connected in a directed and decentralized peer-to-peer setting. The peer-to-peer connectivity is represented by a directed graph $\mathcal{G} = ([n], E)$ where $[n]$ represents the nodes and $E$ represents the edges. The presence of an edge $(i,j) \in E$ indicates the presence of a communication link from node $i$ to node $j$. Please note that since our connectivity graph is directed, the existence of an edge $(i,j) \in E $ does not imply $(j,i) \in E$. Further,  $\mathcal{N}(i)$ denotes the neighbors of node $i$ and is composed of two subsets, in-neighbours $\mathcal{N}_{i}^{in}$ and out-neighbours $\mathcal{N}_{i}^{out}$. In-neighbours $\mathcal{N}_{i}^{in}$ denote the set of nodes with an in-coming edge into node $i$, and out-neighbours $\mathcal{N}_{i}^{out}$ denote the set of nodes with an outgoing edge from node $i$. Additionally, we assume that the graph $\mathcal{G}$ is strongly-connected and has self-loops i.e. each node $i$ is both an in-neighbour and out-neighbour of itself. Also, the graph $\mathcal{G}$ is associated with a mixing matrix $W \in [0,1]^{n \times n}$ such that $w_{ij}$ is the weight from node $i$ to node $j$ ( $w_{ij}$=0 if node $i$ and $j$ are disconnected). The weight matrix $W$ is assumed to be column stochastic (i.e. sum of weights of each column equals to one) with non-negative entries. Further, each node has its own local data coming from distribution $D_i$ and an initial model estimate $x_0^{(i)}$. Our goal is to optimize the global loss function $f(x)$ distributed across the nodes as shown in equation ~\ref{eq:1}.
\begin{equation}
\label{eq:1}
\begin{split}
    \min \limits_{x \in \mathbb{R}^N} f(x) &= \frac{1}{n}\sum_{i=1}^n f_i(x), \\
    where \hspace{2mm} f_i(x) &= \mathbb{E}_{\xi^{(i)} \in D_i}F_i(x^{(i)}, \xi^{(i)}) \hspace{2mm} \forall i \in [n]
\end{split}
\end{equation}
\begin{algorithm}[ht!]
\label{alg:1}
\textbf{Input:} On each node $i\in [n]$ - initialize model parameters $x_0^{(i)}$, $z_0^{(i)}$ = $x_0^{(i)}$, initialize error $\delta_0^{(i)}$ to $0$s and bias weight $u_0^{(i)}$ to $1$, learning rate $\gamma$, averaging rate $\eta$, mixing matrix $W$, and compression operator $C$.\\

1. \textbf{for} $t=0,\hdots,T-1$, at each node $i$, \textbf{do} \\
   \hspace*{6mm} (in parallel for all workers $i \in [n]$)\\
2.\hspace{3mm} Randomly sample minibatch $\xi_t^{(i)}$ from local\\ 
   \hspace*{6mm} distribution $D_{i}$\\
3.\hspace{3mm} Compute local gradient $g_t^{(i)} = \nabla F_i(z_t^{(i)}, \xi_t^{(i)}) $\\
4.\hspace{3mm} Do the local update $\hat{x}_{t}^{(i)} = x_t^{(i)}-\gamma g_t^{(i)}$\\
5.\hspace{3mm} Compute error compensated update $v_{t}^{(i)}=\hat{x}_{t}^{(i)}+\delta_{t}^{(i)}$\\
6.\hspace{3mm} Compress the error compensated variable $C[v_{t}^{(i)}]$\\
   \hspace*{6mm} and update error $\delta_{t}^{(i)} = v_t^{(i)} - C[v_{t}^{(i)}]$\\ 
7.\hspace{3mm} Send compressed variable $C[v_{t}^{(i)}]$ and bias weight $u_t^{(i)}$ 
   \hspace*{6mm} to the out-neighbours of $i^{th}$ node (i.e. $\mathcal{N}_{i}^{out}$) \\
8.\hspace{3mm} Receive $C[v_{t}^{(j)}]$ and $u_t^{(j)}$ from all the in-neighbours\\  
   \hspace*{6mm} of $i^{th}$ node ($j \in \mathcal{N}_{i}^{in}$)\\
9.\hspace{3mm} Do the gossip update for local models and bias weights:
   \hspace*{6mm} $x_{t+1}^{(i)} = \hat{x}_{t}^{(i)}+ \eta \sum_{j\in \mathcal{N}_i^{in}}(W_{ij}-I_{ij})C[v_{t}^{(j)}]$\\
   \hspace*{6mm} $u_{t+1}^{(i)} = u_{t}^{(i)}+ \eta \sum_{j\in \mathcal{N}_i^{in}}(W_{ij}-I_{ij})u_{t}^{(j)}$\\
10.\hspace{3mm}De-bias the updated model: $z_{t+1}^{(i)} = \frac{x_{t+1}^{(i)}}{u_{t+1}^{(i)}}$\\
11. \textbf{end for}\\
12. \textbf{Output Models: $x_{T}^{(i)} = z_{T}^{(i)} \hspace{1mm} \forall i \in [n]$}

\caption{Sparse-Push (SP)}
\end{algorithm}

\emph{Proposed Solution:} Algorithm $1$ describes our proposed solution in a step-by-step manner. The proposed solution consists of three key phases: (a) \emph{Local Update Phase:} Each node $i$ randomly samples a mini-batch of data from its local data distribution and updates its model using local gradient information (Steps 2-4) (b) \emph{Communication Phase:} Calculate and exchange error compensated and compressed model parameters amongst neighbouring nodes (Steps 5-8) (c) \emph{Gossip Update Phase:} Each node updates its local model by performing a gossip update, i.e. a neighbourhood weighted averaging (Steps 9-10).

It may be noted, that the above algorithm simply reduces to vanilla D-PSGD \cite{NIPS2017} algorithm if we assume the compression operator $C$ to be an identity operator, the bias term $u^{(i)}_t$ to be identically one, and weight matrix to be a symmetric and doubly stochastic. In this case, the algorithm is known to converge to a local minima with the same asymptotic rate as centralized stochastic gradient descent. The error-compensated compression scheme using compression operator $C$ augments the vanilla D-PSGD algorithm and enables us to reduce the amount of communication amongst the nodes. Further, the addition of a bias term $u^{(i)}_t$ and de-biasing the updated model at each time step helps extend the algorithm to directed and time-varying graphs\cite{pushsum}. The sections below explain our error-compensated compression scheme, and push-sum protocol to extend our algorithm to directed time-varying graph structures in more detail. 

\emph{Compression Methodology:} Our proposed error-compensation scheme is similar to that in \cite{deepsqueeze} and comprises of two steps. First, we calculate an error compensated update $v_t^{(i)}$ by adding the error introduced by model compression at time step $(t-1)$ to our current model estimate. Subsequently, we communicate the compressed version of error-compensated update, $C[v_t^{(i)}]$ and update our model error parameter. It may be noted that since the error introduced by compressing the model is added back to our model estimate in the next time step, the proposed algorithm essentially behaves like a first-order delta-sigma loop \cite{deltasigma} which is known to converge as long as the local model $\hat{x}_{t}^{(i)}$ does not change very quickly. The learning rate $\gamma$ and averaging rate $\eta$ allow us to easily control the rate of model updates, and hence control convergence for various compression operators $C$. We reiterate, that unlike \cite{koloskova20}, the proposed method does not need to explicitly track and store the model difference with its neighbours, making the method more suitable for time-varying graph topologies.

\emph{Extension to Directed and Time Varying Structures:} To enable our proposed algorithm to work with directed (i.e. asymmetric and column stochastic weight matrix) and time varying graph structures, we propose to use push-sum \cite{pushsum} protocol in our gossip update phase. To better illustrate the need and workings of push-sum algorithm, let us consider the stand alone gossip averaging problem where the goal is to converge each node's final value to be the average of initial values at each node, i.e., at convergence each node converges to $\overline{X} = \frac{1}{n}\sum_{j=1}^n x_{i}$ where $x_{i}$ denotes the initial value at each node $i$. In gossip averaging, this is achieved through repeated neighbour weighted average of the parameters i.e., $x_{t+1}^{(i)}=\sum_{j=1}^n w_{ij}x_{t}^{(j)}$ or in vector form $X_{t+1}=WX_t$. Therefore, at convergence, $X_{\infty} = \lim_{K\to\infty}\prod_{k=0}^{K}W^{(k)}X_0$. Now, leveraging theory of Markov Chains \cite{senata}, it can be shown that if $W^{(k)}$ is column stochastic, then under mild conditions such as strongly connected graph, $\lim_{K\to\infty}\prod_{k=0}^{K}W^{(k)} = \pi\mathbf{1^T}$ where $\pi$ is the ergodic limit of the markov chain and $\mathbf{1}$ is a vector of all ones. Therefore, $X_\infty = \pi(\mathbf{1^T}X_0)$. Further, if the weight matrix $W^{(k)}$ is symmetric as well, i.e. is doubly stochastic, the ergodic limit $\pi$ is simply equal to $1/n$ and simple gossip averaging, i.e. repeated neighbourhood weighted averaging is sufficient to reach convergence. However, when our matrix $W$ is not symmetric, such as in directed graph topologies, knowing the ergodic limit $\pi$ apriori is extremely difficult. To overcome this limitation we augment each node to maintain an additional scalar bias term $u_i$. The scalar bias term is initialized with value of $1$ and is also updated at each time step using the same weight matrix $W$, i.e. $u_{t+1}^{(i)}=\sum_{j=1}^n w_{ij}u_{t}^{(j)}$. This allows the scalar bias term to converge to $U_\infty = \pi\mathbf{1^T1} = \pi$, which can be used to de-bias our estimate by simply calculating $X_\infty/U_\infty{}$. Based on the above, to ensure that our proposed algorithm (Algorithm 1) works with asymmetric column stochastic matrices induced by directed and time varying graphs as well, we introduce an additional bias term $u_i$ for each node $i$, which is updated at each time step in a manner identical to the model weight updates. The updated bias term is then used to de-bias the updated model (Steps 9-10 of our Algorithm 1). 

\section{Experiments}
We evaluate the performance of the proposed algorithm on various time-varying four node directed ring topologies with asymmetric column stochastic mixing matrices. In particular, we test the ability of our algorithm to learn various commonly used deep learning models for image classification task such as ResNet-20, ResNet-110, and VGG models on CIFAR-10 and CIFAR-100 datasets.  In this section, we assume that training dataset is randomly distributed amongst the worker nodes in an IID fashion. Further, we measure the performance of our algorithm in terms of the average test-set accuracy measured across all the worker nodes. We also report the average parameter divergence between the models at the end of the training to ensure the model convergence. The average parameter divergence is computed as the averaged $L_2$ distance been the local model and the averaged model i.e. $\frac{1}{n}\sum_{i=1}^n (||x_i-\Bar{x}||_2)$ \cite{koloskova20} where $n$ is number of nodes, and $\Bar{x} = \frac{1}{n}\sum_{i=1}^n x_i$. Further, similar to \cite{koloskova20},  we evaluate the performance of our algorithm on different model compression operators $C$ such as top-K sparsification. The complete implementation details, results and details on various time-varying directed ring-topologies and additional experimental results are included in the supplementary material.
%\subsection{Implementation details}
%All the experiments were implemented using PyTorch \cite{pytorch}. All local model updates were done using SGD with momentum and weight decay parameters set at 0.9 and $5e^{-4}$. We used an initial learning rate of 0.1 (0.05 for VGG-11) and reduced the learning rate by a factor of 10 after epoch numbers 100 and 150. Our training ran for 200 epochs, with a per node batch size of 32, resulting in an effective batch size of 128 over the four nodes. It may be noted that we tune the averaging rate $\eta$ together with the compression operator $C$. As a rule of thumb, a lower averaging rate was used for higher compression ratios. For instance, we kept averaging rate to be 1 for experiments with no compression, and $1e^{-3}$ for top-$0.1\%$ sparsification. 
\begin{table*}[ht!]
\centering
\caption{ \centering Performance for ResNet models trained with Sparse-Push on CIFAR-10 over directed four node ring topology, with varying levels of top-K sparsification. Compression factor measures the reduction in communication of SP as compared to no compression.}
\label{tab:1}
\begin{tabular}{|l|ccccc|}
\hline
Model & Top-K ($\%$) sparsification & Avg. test  & Data transfer (MB) & Compression  & Parameter  \\
 &  & accuracy & per Iteration & Factor & Divergence\\
 \hline
           & $100$ & $92.09$ & $1.07$ & $0\times$  & $1.968e^{-8}$ \\
           %& $50$   & $91.73$ & $0.812$ & $1.33\times$ &  $2.753e^{-6}$\\
 ResNet-20 & $10$ & $91.27$ & $0.162$  & $6.68\times$ &  $3.408e^{-6}$\\
           & $1$  & $90.68$ & $0.016$ & $66.8\times$ &  $1.512e^{-5}$\\
           & $0.1$  &  $90.51$ &  $0.0018$ & $665\times$ &  $2.845e^{-5}$\\
           & $0$   &  $85.44$ & $0$ &  -   &   $2.0e^{-4}$\\
           \hline
           
           & $100$ & $94.14$ & $6.935$ & $0\times$  & $3.98e^{-9}$ \\
        %  & $10$ & $92.64$ & $1.036$  & $6.69\times$ &  $2.71e^{-7}$\\
 ResNet-110 & $1$  & $93.32$ & $0.104$ & $66.4\times$ &  $9.97e^{-7}$\\
           & $0.1$  &  $93.26$ &  $0.011$ & $611\times$ &  $1.85e^{-6}$\\
           & $0$   &  $86.38$ & $0$ &  -   &   $1.57e^{-5}$\\
           
 \hline
       & $100$ & $90.86$ & $37.95$ & $0\times$  & $3.48e^{-10}$ \\
    & $10$ & $90.32$ & $5.69$  & $6.66\times$ &  $4.57e^{-8}$\\
    VGG11 & $1$  & $90.22$ & $0.569$ & $66.7\times$ &  $2.3e^{-7}$\\
          & $0.1$  &  $88.81$ &  $0.057$  & $667\times$ &  $5.1e^{-7}$\\
          & $0$   &  $80.90$ & $0$ &  -   &   $3.76e^{-6}$\\
  \hline
 \end{tabular}
\end{table*}

\subsection{Results}

Table.~\ref{tab:1} summarizes the performance of our proposed algorithm on ResNet models for CIFAR-10 dataset over a four-node directed ring topology with varying levels of communication compression (using top-K sparsification). 
We demonstrate upto $67\times$ reduction in communication while trading off $1.4\%$ accuracy and $665\times$ compression for $1.6\%$ accuracy loss. Further, we evaluate our algorithm on various time-varying directed graph topologies and achieve similar performance (Please refer supplementary section. A3 and table. A2).
%Our proposed SP algorithm allows for training ResNet-20 models over directed and time-varying graph topologies with $67\times$ reduction in communication while trading off $1.4\%$ in the test accuracy and $665\times$ compression for $1.6\%$ loss in test accuracy.
\begin{table}[ht!]
\centering
\caption{ \centering Test accuracy for ResNet-20 with CIFAR-10 dataset across different decentralized training algorithms with top-1\% sparsification}
\label{tab:c2}
\begin{small}
\begin{tabular}{|l|cccc|}
\hline
Method & CHOCO & Deep & SP & SCSP\\
 & -SGD & -Squeeze &  & \\
\hline
    Test-accuracy  & $91.73$& $90.47$& $90.68$ & $91.27$\\
 \hline
 \end{tabular}
 \end{small}
\end{table}
We validate the performance of our approach on models with varying network capacity and across different datasets (Table.\ref{tab:3} Column 4). We also report the results with K=0 top-K sparsification (i.e. no communication amongst nodes) to validate that at-least some minimal amount of communication is necessary for good generalization. Further, table.~\ref{tab:c2} compares test accuracy of our proposed SP and SCSP (Section 5) algorithms with two state-of-the art algorithms, Choco-SGD and Deep-Squeeze. To ensure a fair comparison we report results using ring-topology graphs with same spectral gap. As evident, all three models achieve similar test-accuracy. However, unlike others, our approach also supports time-varying and directed graphs which further leads to a $2\times$ reduction in inter-node communication.

\section{Skew-Compensated Sparse-Push (SCSP)}
This section discusses the effect of non-IID data distribution amongst the worker nodes during communication efficient decentralized training over time-varying directed peer-to-peer graphs. In particular, we highlight the interplay between communication compression and amount of non-IIDness in data distribution amongst the worker nodes. Further, we propose a simple communication efficient strategy that helps us reduce the impact of non-IIDness. Additionally, we highlight how the proposed approach helps improve performance of our Sparse-push algorithm even in IID data settings.
\begin{figure}[ht!]
%\vskip 0.01in
\begin{center}
\centerline{\includegraphics[width=0.9\columnwidth,scale=0.1]{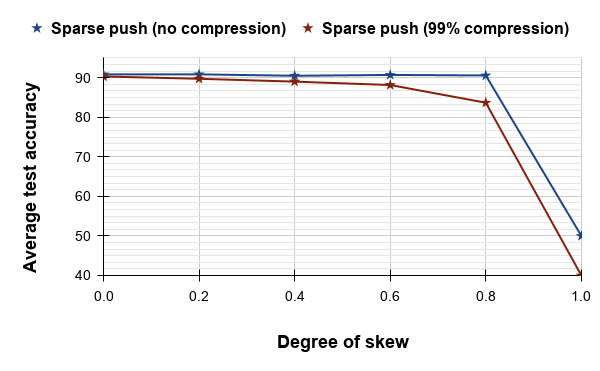}}
\caption{Test set accuracy vs data skew for VGG11 model trained on CIFAR-10 dataset using sparse-push algorithm with and without communication compression}
\label{fig:skew}
\end{center}
\vskip -0.4in
\end{figure}

To understand the impact of non-IIDness we start by defining \emph{data-skew} as a measure of non-IIDness in data distribution amongst the worker nodes. We define Data-skew, within the context of image classification datasets such as CIFAR-10/CIFAR-100, as the fraction of training data that is partitioned amongst the worker nodes in a non-IID manner. For instance, a data-skew of $0.8$ indicates $80\%$ of the training data is partitioned amongst the workers in a non-IID manner and the remaining $20\%$ data is distributed in an IID-manner. To generate non-IID training sets, we use skewed label partitions \cite{hsieh20a}.
%we sort the available training data based on its labels and sequentially assign the sorted data amongst the worker nodes. 
For instance, let us consider a dataset of 100 examples containing 10 examples each from 10 different classes. To generate a dataset with skew of $0.8$ amongst 4 nodes, we first randomly sample $80\%$ of each class from the training data and then sort the sampled data as per its label. First 20 examples from the sorted data are assigned to node 1, the next 20 to node 2 and so on. The remaining $20\%$ of the data is randomly distributed across the 4 nodes.

Figure.~\ref{fig:skew} shows the performance of sparse push algorithm on VGG11 trained on CIFAR10 dataset with varying degrees of data skew over a four-node directed ring topology for two cases - Network is trained with (a) no communication compression, and (b) using top-K sparsification with K=1 (i.e. 99\% compression). It may be noted that while the two models behave almost identically at zero data-skew, for larger data-skew values the test-set accuracy degrades significantly more for models that are trained with 99\% communication compression. This is not very surprising because increasing the data-skew (in training data) amongst the nodes increases the difference among the models being learnt by different worker nodes, and hence, they would need to communicate more to overcome this difference and converge to a common solution.
\begin{algorithm}[ht!]
\textbf{Input:} Same as Algorithm 1\\

1. \textbf{for $t=0,\hdots,T-1$ do} \\
  \hspace*{6mm} (in parallel for all workers $i \in [n]$)\\
2.\hspace{3mm} follow the steps 2-9 from Algorithm 1\\
3. \textbf{end for}\\
4. \textbf{for $t=T,\hdots,T+k$ where $(k<<T)$ do} \\
%. \hspace{3mm} (in parallel for all workers $i \in [n]$)\\
5.\hspace{3mm} Send the model parameters $x_{t}^{(i)}$ and bias weight $u_t^{(i)}$ to  
  \hspace*{6mm} the out-neighbours of $i^{th}$ node (i.e. $N_{(i)}^{out}$) \\
6.\hspace{3mm} Receive $x_{t}^{(j)}$ and $u_t^{(j)}$ from all the in-neighbours of $i^{th}$\\ 
  \hspace*{6mm} node ($j \in N_{in}^{(i)}$)\\
7.\hspace{3mm} Do the gossip update for local models and bias weights:
  \hspace*{6mm} $x_{t+1}^{(i)} = {x}_{t}^{(i)}+ \eta \sum_{j\in N_{in}^{(i)}}(W_{ij}-I_{ij})x_{t}^{(j)}$\\
  \hspace*{6mm} $u_{t+1}^{(i)} = u_{t}^{(i)}+ \eta \sum_{j\in N_{in}^{(i)}}(W_{ij}-I_{ij})u_{t}^{(j)}$\\
8.\hspace{3mm} De-bias the updated model: $z_{t+1}^{(i)} = \frac{x_{t+1}^{(i)}}{u_{t+1}^{(i)}}$\\
9. \textbf{end for}\\
10. \textbf{Output Models: $x_{T+k}^{(i)} = z_{T+k}^{(i)} \hspace{1mm} \forall i \in [n]$}
\caption{Skew-Compensated Sparse Push (SCSP)}
\end{algorithm}

\begin{table*}[ht!]
\centering
\caption{ \centering Performance of Sparse-Push (SP) and Skew-Compensated Sparse Push (SCSP) with varying levels of top-K sparsification on a four node directed ring graph. Compression factor measures the reduction in communication of SCSP as compared to no compression.}
\label{tab:3}
\begin{tabular}{|l|ccccc|}
\hline
Dataset/ & top-K$\%$  & Compression & Avg. test accuracy $(\%)$ & Avg. test accuracy $(\%$) &  Accuracy gain $(\%$) \\
 Network & Sparsification & factor & with SP & with SCSP & from SCSP  \\
\hline
  CIFAR-10 & $100$ & $0\times$  & $92.09$ & $92.09$  & $-$ \\
    %   CIFAR-10   & $10$ & $6.64\times$ & $91.27$ &  $91.47$ & $0.20$  \\
 ResNet-20 & $1$  &  $64.3\times$ & $90.68$ & $91.27$  & $0.59$ \\
           & $0.1$  &  $466\times$ &  $90.51$ & $91.09$ & $0.58$ \\
           
\hline
       CIFAR-10      & $100$ & $0\times$  & $91.20$ & $91.20$ & $-$ \\
   %    CIFAR-10   & $10$ & $6.50\times$ & $90.49$ & $90.58$  &   $0.09$\\
 VGG-11 & $1$  &  $63.9\times$ & $90.22$ & $90.49$  &  $0.27$\\
           & $0.1$  &  $467\times$ &  $88.81$ &  $89.99$ &  $1.18$\\
 \hline
   CIFAR-100 & $100$ & $0\times$ & $72.10$ & $72.10$   & $-$ \\
%CIFAR-100   & $10$ & $6.64\times$ & $71.17$ & $71.32$   &  $0.15$\\
ResNet-110 & $1$  & $63.6\times$ & $69.98$ & $71.28$ &   $1.30$\\
           & $0.1$  &  $438\times$ & $68.25$ &  $69.87$ &   $1.62$\\
          
  \hline
 \end{tabular}
\end{table*}

Based on the above, the key question is, \textit{``Given the trade-off between data-skew and communication compression, how should we communicate such that we maximize the joint task performance while minimizing the amount of communication among the nodes even in presence of data-skews?}" One potential approach is to formulate the above as an explicit optimization problem, however, repeatedly solving the optimization problem on energy constrained edge device with time-varying graphs may not be trivial. Therefore, instead, we propose a simple heuristic approach based on our experimental observations, ``We communicate amongst the nodes with a large compression ratio (say 99\% or 99.9\% compression) for most part of training, and towards the end of training we communicate with no compression". This heuristic is based on our observation that full communication is far more vital towards the end of the training when the models are already partially trained. Formally, Algorithm 2-Skew Compensated Sparse Push (SCSP), depicts our proposed approach: Train the model using Sparse-Push algorithm with a high compression ratio for the first $T$ time-steps, followed by $k << T$ time-steps of gossip averaging with no compression. 

\begin{figure}[ht!]
\begin{center}
\centerline{\includegraphics[width=\columnwidth]{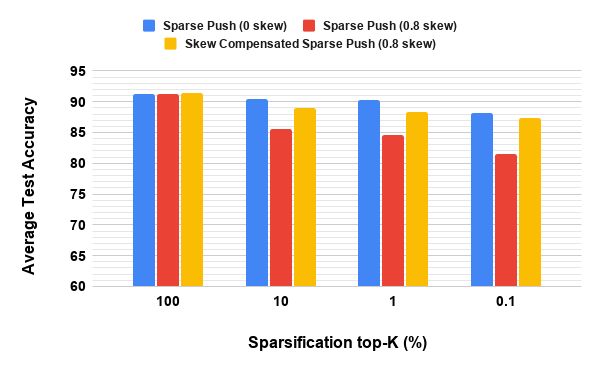}}
\caption{Comparison of test set accuracy of VGG11 trained on a four-node directed ring topology using SP and SCSP algorithms with varying degree of communication compression on IID-data (i.e. 0-Skew) and non-IID ($0.8$ data-skew) CIFAR-10 dataset.}
\label{fig:vgg}
\end{center}
\vskip -0.2in
\end{figure}

We evaluate the performance of SCSP, by training VGG11 model on a four-node directed ring topology on non-IID distributed CIFAR-10 training data with varying degrees of data-skew. Figure.\ref{fig:vgg} shows the performance of our approach and depicts the test accuracy achieved when VGG11 model is trained with SCSP with varying degree of communication compression using IID-data (i.e. 0-Skew) and non-IID data with $0.8$ data-skew. As evident, in-case of no communication compression we achieve identical test set accuracy between models trained using Sparse push (SP) and SCSP for both IID and non-IID datasets. However, as we increase the amount of communication compression, the test set accuracy significantly drops for models trained using SP on non-IID dataset with $0.8$ skew. For instance, for $99.9\%$ compression scenario, SP trained models loose $~7\%$ test set accuracy when trained using non-IID dataset (compared to IID-datset). However, leveraging SCSP, we are able recover most of this performance drop and the difference in test-set accuracy between models trained with SP on IID-dataset, and SCSP on non-IID datasets is less than ~$0.5\%$ for top-$0.1\%$ case. It may be noted that for the above experiments we chose $k$, the number of iterations to run full-communication gossip averaging for, to be 40, which is extremely small compared to number of iterations $T$ (=200 epochs $\times$ 391 iterations per epoch), thus keeping the communication overhead of SCSP to be extremely small. 

Further, we evaluate, if SCSP algorithm can help improve performance for IID datasets as well. Table $4$ summarizes our results for ResNet and VGG models and illustrates the ability of SCSP to boost performance by upto $1.6\%$, especially in settings with large communication compression. \textbf{We highlight, our proposed SCSP algorithm achieves upto 438x reduction in communication over a directed time-varying graph topology with $\mathbf{1\%}$ degradation in performance on ResNet-20 and VGG11 models over CIFAR-10 dataset. Notably, we achieve an order of magnitude higher communication compression, support directed and time-varying graphs, while maintaining similar test-accuracy as the current state-of-the-art algorithms such as CHOCO-SGD and Deep-Squeeze.}

\section{Hardware and Energy Implications of Our Proposed Approach}
\begin{figure}[ht!]
\begin{center}
\centerline{\includegraphics[width=\columnwidth]{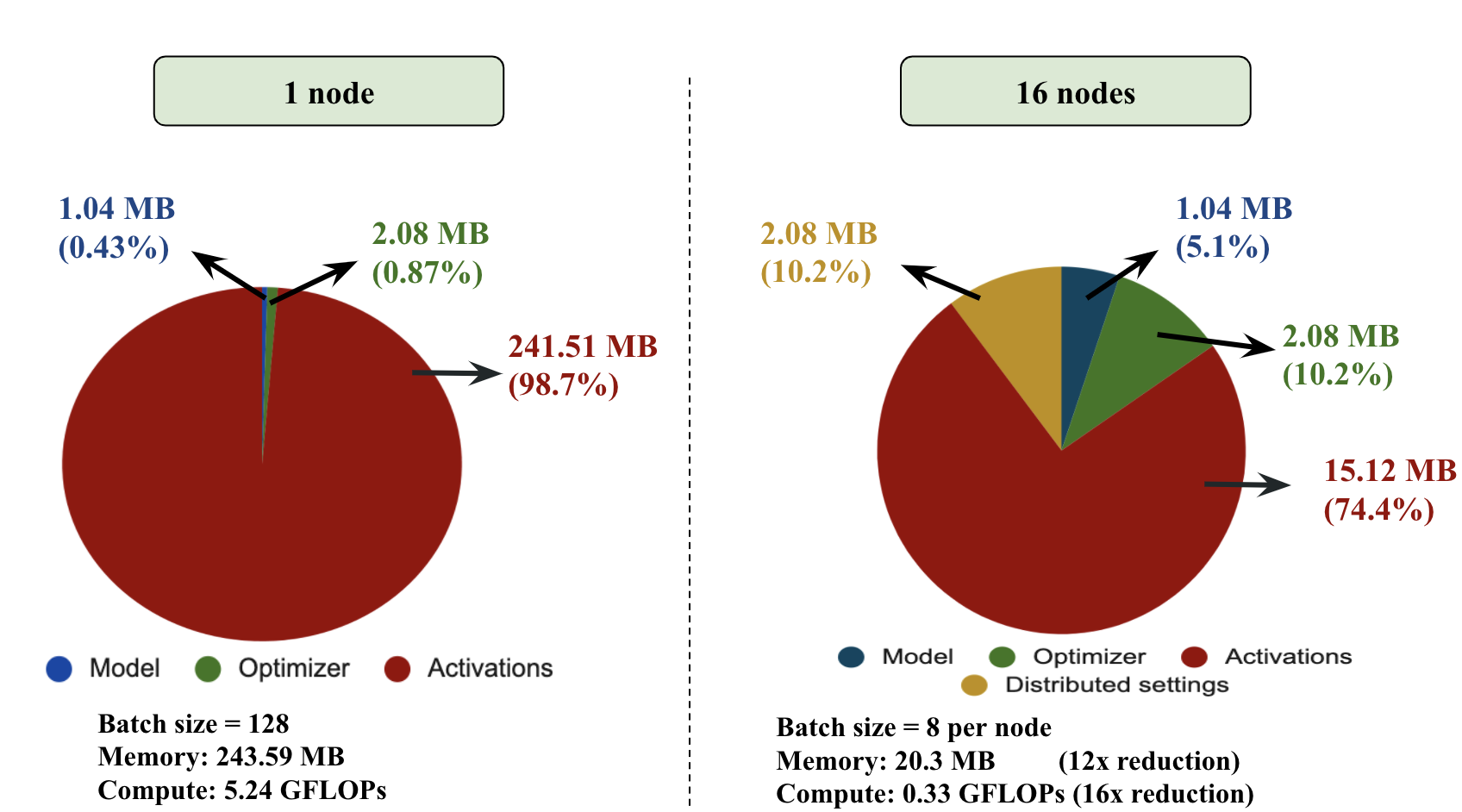}}
\caption{Memory and compute requirements for training ResNet-20 with CIFAR-10 dataset on a single node vs sixteen nodes connected in a peer-to-peer network}
\label{fig:mprof}
\end{center}
\vskip -0.1in
\end{figure}

This section highlights how our proposed algorithm, SCSP, can help enable energy efficient on-device training as well. In particular, we note that communication efficient decentralized distributed learning over peer-to-peer wireless devices significantly reduces per device memory and compute requirements, helping enable training on edge devices such as smartphones and drones. Figure \ref{fig:mprof}, illustrates per device memory and compute requirements for training a ResNet-20 model over CIFAR-10 dataset using a single device or a directed ring network of sixteen devices while maintaining the same equivalent batch size. 
We highlight that distributed learning over sixteen nodes helps reduce the per device memory requirement from 243MB to 20.3MB, and compute requirement from 5.24GFlops to 0.33GFlops. This large reduction in per device memory requirements can be attributed to the fact that most of the memory during training is occupied by activation memory, i.e. the memory needed to store activations, gradients, and errors. However, activation memory is not fundamental and is purely a function of batch size. Distributed and decentralized learning allows us to distribute our data over multiple devices, thus reducing the required batch size per device, while still keeping a constant throughput. Notably, this approach introduces additional constraints such as communication overhead, ability to deal with time-varying and directed graphs, and an ability to learn with non-IID datasets amongst worker nodes, exactly the challenges that our proposed SCSP training algorithm tries to solve. 

\section{Conclusion and Future Work}
In this work we proposed, \emph{Sparse-Push}, a communication efficient decentralized distributed training algorithm that supports training over peer-to-peer, directed and time-varying graph topologies. Our proposed algorithm demonstrated upto 438$\times$ reduction in communication with only $1\%$ degradation in performance when training ResNet-20 and VGG11 models over CIFAR-10 dataset. We also demonstrated the scalability of our approach to larger datasets such as CIFAR-100. Further, we demonstrated the interplay between communication compression and non-IID datasets in decentralized distributed learning settings and highlight how communication compression can lead to drop in performance over non-IID distributed data. We emphasize the need for training algorithms, that jointly optimize communication efficiency and performance over non-IID datasets, and as a first step towards that goal, we proposed  \emph{Skew-Compensated Sparse Push} algorithm that helps recover the performance drop caused due to non-IID datasets while maintaining similar levels of communication compression. As next steps, we plan to extend this work to other tasks and models beyond image classification, for instance, on speech recognition and translation using transformer models and robotic perception and manipulation using reinforcement learning models. Further, our current work necessitates synchronous updates, and we plan to extend our proposed algorithms to allow for asynchronous updates to ease real-life deployment of our algorithm on edge devices such as smartphones and drones. 

\bibliography{example_paper}
\bibliographystyle{icml2020}

 \newpage

\section{Supplementary Material}
\subsection{Assumptions}
We highlight the various assumptions in our proposed Sparse-Push and Skew-Compensated sparse-push algorithms. 
\begin{enumerate}
    \item \textbf{Graph Structure:} The connectivity graph of workers ($G$) is strongly connected.
    \item \textbf{Column stochastic mixing matrix:} The mixing matrix $W$ is a real column stochastic matrix i.e. $\mathbf{1}^T W=W$, where $\mathbf{1}$ is the vector of all $1's$.
    \item \textbf{Lipschitz Gradients:} Local loss functions $f_i(.)$ has L-lipschitz gradients for all $i \in [1,n]$ i.e.\\
    $||f_i(x)-f_i(y)|| \leq L ||x-y|| \hspace{3mm} \forall x,y \in \mathbb{R}^d$ 
    \item \textbf{Bounded Variance:} The variance of the stochastic gradients are assumed to be bounded.\\
    $\mathbb{E}_{ \xi \sim D_i} || \nabla F_i(x; \xi) - \nabla f_i(x)||^2 \leq \sigma^2,$  (Inner variance).\\
    $ \frac{1}{n} \sum_{i=1}^n || \nabla f_i(x) - \nabla f(x)||^2 \leq \zeta^2 \hspace{2mm} \forall i,x$ (Outer variance).
    \item \textbf{Bounded signal-to-noise ratio:} The magnitude of the compression error is bounded by the magnitude of the input vector.\\
    $\mathbb{E}||C[x]-x||^2 \leq \alpha^2 ||x||^2, \forall x$ where $\alpha \in [0,1)$ and $C$ is the compression operator. 
    \item \textbf{Initialization:} The model parameters on all the nodes are initialized to the same random values.
\end{enumerate}

It may be noted that assumptions 1, 3-6 are similar to that of deep-squeeze algorithm \cite{deepsqueeze} and are commonly used in most decentralized distributed training algorithms with communication compression. We reiterate, assumption 2 replaces the commonly used doubly stochastic mixing matrix requirement with a column stochastic mixing matrix, allowing us to extend our work to directed graph topologies. 
 \begin{table}[ht!]
\centering
\caption{ \centering Averaging rate $\eta$ used in our experiments for different compression/sparsification factors}
\label{tab:eta}
\begin{small}
\begin{tabular}{c|c}
\hline
Top-K ($\%$) sparsification & Averaging rate $\eta$ \\
\hline
   100 & 1\\
   50 & 0.08\\
   10 & 0.01\\
   1 & 0.005\\
   0.1 & 0.001\\
 \hline
 \end{tabular}
 \end{small}
\end{table}
\begin{figure}[ht!]
\begin{center}
\centerline{\includegraphics[width=\columnwidth]{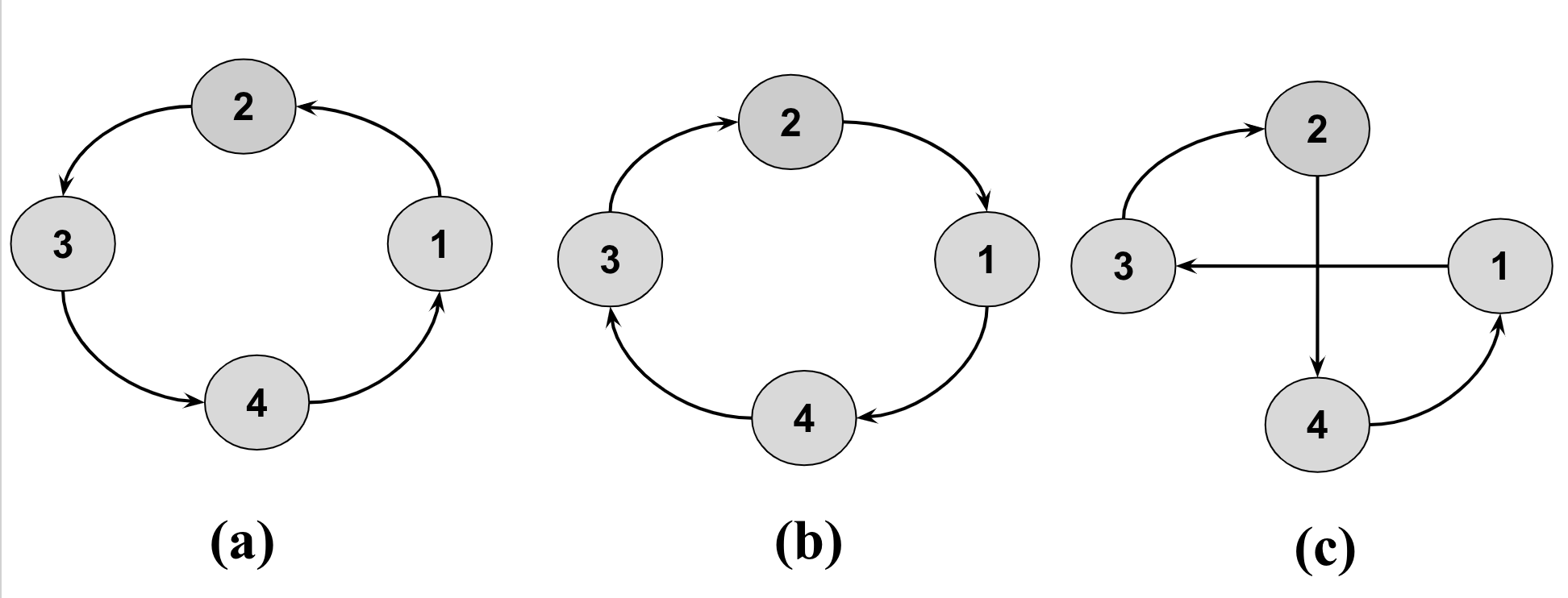}}
\caption{The graph topologies (of 4 nodes) used in our experiments. For directed graph experiments we have used the ring topology (a). For experiments with time-varying directed graph, we rotate between the graphs (a), (b) and (c) after every $n^{th}$ epoch.}
\label{fig:graphs}
\end{center}
\vskip -0.2in
\end{figure}

\subsection{Implementation details}
 To show the performance and scalability of the proposed algorithm, we trained various standard architectures such as ResNet and VGG11 (without batch normalization layers) on bench-marked image recognition datasets such as \href{https://www.cs.toronto.edu/~kriz/cifar.html}{CIFAR-10 and CIFAR-100} using our proposed SP, and SCSP algorithms. 
All the experiments were implemented using PyTorch \cite{pytorch} and were conducted on a system with 4 \textit{Nvidia GTX 2080ti} GPUs.
All local model updates were done using SGD with momentum and weight decay 0.9 and $1e^{-4}$ ($5e^{-4}$ for VGG). 
For ResNet architectures, we used an initial learning rate of 0.1 and reduced the learning rate by a factor of 10 after epoch numbers 100 and 150. For VGG11 network, we used an initial learning of 0.05 and reduced it by a factor of 2 after every 30 epochs. We trained our networks for a total of 200 epochs with a per node batch size of 32, resulting in an effective batch size of 128 over the four nodes. 
The input images to the model are normalized using channel means and standard deviations.
The following transforms were applied to the training data to get the benefit of data augmentation: \textit{transforms.RandomCrop} and \textit{transforms.RandomHorizontalFlip}.

\begin{figure*}[ht!]
\begin{center}
\begin{small}
\[ \arraycolsep=0.8\arraycolsep\ensuremath{\begin{bmatrix}
1-\frac{1}{2}\eta & 0 & 0 & \frac{1}{3}\eta  \\
\frac{1}{2}\eta & 1-\frac{1}{3}\eta & 0 & 0  \\
0 & \frac{1}{3}\eta & 1-\frac{1}{2}\eta & 0  \\
0 & 0 & \frac{1}{2}\eta & 1-\frac{1}{3}\eta   \nonumber
\end{bmatrix} \hspace{3mm}
\begin{bmatrix}
1-\frac{1}{2}\eta & \frac{1}{3}\eta& 0 &0  \\
0 & 1-\frac{1}{3}\eta & \frac{1}{2}\eta & 0  \\
0 & 0 & 1-\frac{1}{2}\eta & \frac{1}{3}\eta  \\
\frac{1}{2}\eta & 0 & 0 & 1-\frac{1}{3}\eta   \nonumber
\end{bmatrix} \hspace{3mm}
\begin{bmatrix}
1-\frac{1}{2}\eta & 0 & 0 & \frac{1}{3}\eta  \\
0 & 1-\frac{1}{3}\eta & \frac{1}{2}\eta & 0  \\
\frac{1}{2}\eta & 0 & 1-\frac{1}{2}\eta & 0  \\
0 & \frac{1}{3}\eta & 0 & 1-\frac{1}{3}\eta   \nonumber
\end{bmatrix}} \]
\end{small}
\label{fig:mat}
\end{center}
\caption{ \centering The assymetric and column stochastic mixing matrices used in our experiments corresponding to the graph topologies (a), (b), (c) respectively shown in fig.~\ref{fig:graphs}. Note $\eta$ is the averaging rate which is set according to the compression ratio.}
\end{figure*}

\begin{table*}[ht!]
\centering
\caption{ \centering Performance of Sparse-Push (SP) and Skew-Compensated Sparse Push (SCSP) algorithms with varying levels of top-K sparsification over time varying directed graphs on ResNet-20 architecture, CIFAR10 dataset. The graph structures (a), (b), (c) are shown in fig.~\ref{fig:graphs}}
\label{tab:tv}
\begin{tabular}{|l|ccccc|}
\hline
Graph & varying every&top-K$\%$   & Avg. test accuracy $(\%)$ & Avg. test accuracy $(\%$) &  Accuracy gain $(\%$) \\
 structures & $n^{th}$ epoch & Sparsification & with SP & with SCSP & from SCSP  \\
\hline
          &  & $100$  & $91.92$ & $91.92$  & $-$ \\
 (a), (b) & $11$  &  $1$ & $90.75$  &  $91.61$ & $0.86$ \\
           &   &  $0.1$ &  $90.08$ & $90.74$ & $0.66$ \\
           
\hline
             & & $100$  & $91.63$ & $91.63$ & $-$ \\
 (a), (b), (c) & $6$  &  $1$ &  $91.00$& $91.46$& $0.46$  \\
           &   &  $0.1$ &  $89.92$ &  $90.53$ &  $0.61$\\
 \hline
 
 \end{tabular}
\end{table*}

Further, please note that the averaging rate $\eta$ was tuned together with the compression operator $C$. As a rule of thumb, a lower averaging rate was used for higher compression ratios as shown in table.~\ref{tab:eta}. 
The experiments were conducted on four node topologies with asymmetric and column stochastic mixing matrix as shown in fig.~\ref{fig:graphs}.
All the results presented in the main paper i.e. table. 2-4, fig. 1-2 are conducted on 4 node directed ring topology (a) in fig.~\ref{fig:graphs}. The source code for our experiments will be released shortly. %The source code can be found \href{https://drive.google.com/drive/folders/1vDjjX8_I7ObytUtGy08wd0AIz2Zg4AUF?usp=sharing}{here.}
\begin{figure}[ht!]
%\vskip 0.01in
\begin{center}
\centerline{\includegraphics[width=0.9\columnwidth,scale=0.1]{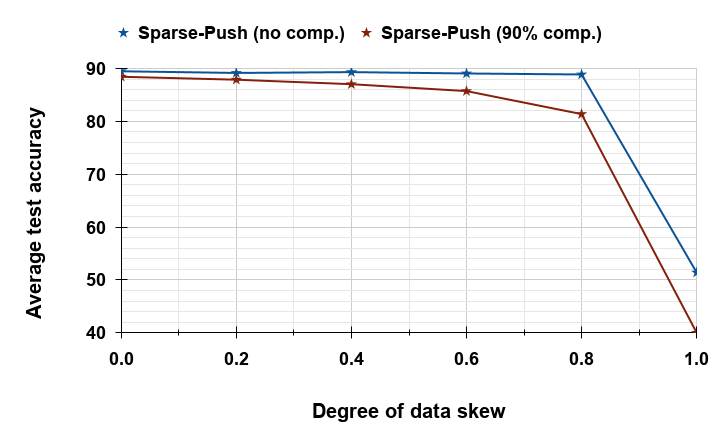}}
\caption{Test set accuracy vs data skew for ResNet-20 model with batch normalization layers replaced by group normalization (group size = 2) trained on CIFAR-10 dataset using sparse-push algorithm with and without communication compression}
\label{fig:skew_res}
\end{center}
\vskip -0.2in
\end{figure}

\begin{table*}[ht!]
\centering
\caption{ \centering Performance of Sparse-Push (SP) and Skew-Compensated Sparse Push (SCSP) algorithms using various quantization compression operators. Compression factor measures the reduction in communication of SCSP as compared to no compression.}
\label{tab:qsp}
\begin{tabular}{|l|ccccc|}
\hline
Dataset/ & Bit  & Compression & Avg. test accuracy $(\%)$ & Avg. test accuracy $(\%$) &  Accuracy gain $(\%$) \\
 Network & precision & factor & with SP & with SCSP & from SCSP  \\
\hline
               & $32$ & $0\times$  & $92.09$ & $92.09$  & $-$ \\
    CIFAR-10   & $4$ & $7.96\times$ & $91.17$ & $91.19$  &  $0.02$ \\
 ResNet-20 & $2$  &  $15.84\times$ & $91.06$ &  $91.15$ & $0.09$ \\
           & $1$  &  $31.36\times$ & $90.87$  & $91.05$ & $0.18$\\
 \hline
 \end{tabular}
\end{table*}

\subsection{Additional Results}
In this section we present additional results using Sparse-Push (SP) and Skew Compensated Sparse-Push (SCSP) algorithms over various time-varying directed graph topologies.  Table.~\ref{tab:tv} shows the performance of SP and SCSP on various time varying directed graph structures. The various time varying graph structures used in our experiments are shown in fig.~\ref{fig:graphs} and the structure of the graph varies every $n^{th}$ epoch. For example, the results in the second row of table.~\ref{tab:tv} corresponds to a time varying directed graph where the graphs structure changes as $[ (a) \rightarrow (b) \rightarrow (c) \rightarrow (a) \rightarrow \hdots ]$ every 6 epochs. We achieve a $466\times$ reduction in communication with $1.0-1.2\%$ loss in accuracy, demonstrating the ability of our algorithm to achieve very similar results between time-varying directed graphs and static directed graphs. 
Further, we also evaluate the performance our proposed algorithms for random quantization compressor operator \cite{alistarh2017qsgd, koloskova19a} by training RESNET-20 over CIFAR10 dataset on a four-node directed ring topology (Table.~\ref{tab:qsp}). We achieve $31.36 \times$ compression when communicating with 1 bit precision using SCSP algorithm while trading off $1\%$ accuracy.
\begin{figure}[ht!]
\begin{center}
\centerline{\includegraphics[width=\columnwidth]{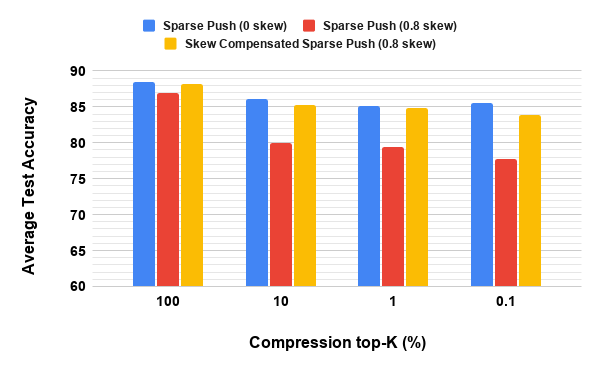}}
\caption{Comparison of test set accuracy of ResNet-20 model with batch normalization layers replaced by group normalization (group size = 2) trained on a four-node directed ring topology using SP and SCSP algorithms with varying degree of communication compression on IID-data (i.e. 0-Skew) and non-IID ($0.8$ data-skew) CIFAR-10 dataset.}
\label{fig:res}
\end{center}
\vskip -0.2in
\end{figure}

We also validate the performance of SCSP algorithm on ResNet architecture over 4-node directed ring topology with non-IID data partitions. Since the batch-normalization layers degrade the performance in the presence of non-IIDness we replace them with group normalization layers as suggested in \cite{hsieh20a}. For the CIFAR-10 dataset on ResNet-20 architecture we use a group size of 2 for all the group normalization layers which gives a baseline accuracy of $88.5\%$ with 0 data skew and no compression. Fig.~\ref{fig:skew_res} shows the degradation of performance of Sparse-Push over CIFAR-10 dataset on ResNet-20 architecture with change in data skew in the presence of communication compression compared to no compression. With a skew of $0.8$ and top-10 sparsification (i.e. $90\%$ compression), the performance degrades by around $6\%$. However, leveraging SCSP, we are able recover most of this performance drop and the difference in test-set accuracy between models trained with SP on IID-dataset, and SCSP on non-IID datasets is less than ~$0.4\%$ for top-$1\%$ case. Thus, we empirically show the convergence of Sparse-Push algorithm and verify the performance improvement for non-IID data using Skew-Compensated Sparse-Push algorithm through our experiments on various architectures and datasets.

\end{document}